\documentclass[sigconf]{acmart}
\usepackage{pifont}
\usepackage{balance}
\newcommand{\cmark}{\ding{51}}
\newcommand{\xmark}{\ding{55}}

\AtBeginDocument{%
  }

\copyrightyear{2026}
\acmYear{2026}
\setcopyright{cc}
\setcctype{by-nc-nd}
\acmConference[MM '26]{Proceedings of the 34th ACM International Conference on Multimedia}{November 10--14, 2026}{Rio de Janeiro, Brazil}
\acmBooktitle{Proceedings of the 34th ACM International Conference on Multimedia (MM '26), November 10--14, 2026, Rio de Janeiro, Brazil}
\acmDOI{10.1145/3767308.3838663}
\acmISBN{979-8-4007-2213-4/2026/11}

\begin{document}

\title{PhysScene: A Scene Graph Dataset for Scientific Visual Reasoning in Physics Experiments}

\author{Minghao Zou}
\affiliation{%
	\institution{Cardiff University}
	\city{Cardiff}
	\country{United Kingdom}
}
\email{zoum1@cardiff.ac.uk}

\author{Qingtian Zeng}
\affiliation{%
	\institution{Shandong University of Science and Technology}
	\city{Qingdao}
	\country{China}
}
\email{qtzeng@sdust.edu.cn}

\author{Shangkun Liu}
\authornote{Corresponding authors.}
\affiliation{%
	\institution{Shandong University of Science and Technology}
	\city{Qingdao}
	\country{China}
}
\email{liushangkun@sdust.edu.cn}

\author{Yanda Meng}
\affiliation{%
	\institution{King Abdullah University of Science and Technology}
	\city{Thuwal}
	\country{Saudi Arabia}
}
\email{yanda.meng@kaust.edu}

\author{Guanghui Yue}
\affiliation{%
	\institution{Shenzhen University}
	\city{Shenzhen}
	\country{China}
}
\email{yueguanghui@szu.edu.cn}

\author{Baoquan Zhao}
\affiliation{%
	\institution{Sun Yat-sen University}
	\city{Zhuhai}
	\country{China}
}
\email{zhaobaoquan@mail.sysu.edu.cn}

\author{Abdulmotaleb El Saddik}
\affiliation{%
	\institution{University of Ottawa}
	\city{Ottawa}
	\country{Canada}
}
\email{elsaddik@uottawa.ca}

\author{Wei Zhou}
\authornotemark[1]
\affiliation{%
	\institution{Cardiff University}
	\city{Cardiff}
	\country{United Kingdom}
}
\email{zhouw26@cardiff.ac.uk}
\renewcommand{\shortauthors}{Minghao Zou et al.}

\begin{abstract}
Scene Graphs (SGs) provide structured representations of visual scenes by modeling objects and their pairwise relationships. Despite recent progress, existing datasets primarily focus on generic natural contexts, leaving domain-specific and function-oriented scenes largely underexplored. This limitation restricts the evaluation of relational reasoning in scientific experimental scenes, thereby hindering the development of intelligent monitoring, analysis, and related applications in such scenes. To address this gap, we introduce PhysScene, the first SG dataset tailored to physics experiments. PhysScene encompasses specialized instruments, structured experimental setups, and functional relations intrinsic to experimental environments, enabling reasoning that extends beyond spatial co-occurrence to logical dependencies. Rather than pursuing large data scale, PhysScene focuses on strong semantic constraints and high relation density in experimental scenes, posing new challenges for existing scene parsing algorithms while offering opportunities for further improvements. Extensive analyses and experiments show that PhysScene complements existing benchmarks and establishes a valuable testbed for advancing scientific visual reasoning. The dataset is publicly available at \href{https://github.com/ZMH-SDUST/PhysScene}{https://github.com/ZMH-SDUST/PhysScene}.
\end{abstract}




\maketitle

\section{Introduction}
\label{sec:intro}

Scene Graph Generation (SGG) aims to represent visual scenes in a structured manner by detecting objects and modeling their pairwise relationships \cite{zhang2023learning}. Such representations support a variety of downstream tasks, such as image retrieval, visual question answering, and logical reasoning \cite{chang2021comprehensive}. Driven by the development of datasets, SGG has achieved substantial progress in recent years.\par

However, existing SGG datasets predominantly focus on generic natural imagery, such as household scenes, street views, and natural environments \cite{krishna2017visual, belz2018spatialvoc2k}. While these datasets are large and diverse, their relational annotations are largely grounded in common, visually apparent interactions. As a result, current benchmarks provide limited support for evaluating SGG models in domain-specific environments where objects, relations, and configurations are tightly coupled with specialized tasks \cite{li2024aug, li2025star, zou2025physlab}.\par

Scientific laboratory settings present a particularly distinctive scenario. Compared with generic natural scenes \cite{yang2019spatialsense, yang2022panoptic}, scientific experiments involve specialized instruments, structured experimental setups, and functional interactions between operators and instruments. Visually similar experimental objects may serve different roles, and relations in experiments often reflect operational intent beyond mere spatial proximity. Detecting domain-specific scene elements and uncovering experimental semantic constraints introduce new challenges beyond conventional object co-occurrence modeling, highlighting the need for specialized benchmarks that can assess relational modeling in experimental contexts. \par

Motivated by prior work on modeling experimental procedures \cite{zou2025physlab}, we take physics laboratory experiments as a representative setting of scientific experimental scenes and construct PhysScene, the first Scene Graph (SG) dataset for such scenes. PhysScene comprises 4.5K high-resolution images with 45.9K annotated object instances, spanning 34 object and 39 relation categories, with a total of 130.4K relational triplets. Each image is annotated under a multi-level scheme \cite{krishna2017visual}, including object categories and bounding boxes, object-level attributes, and pairwise relationships between objects such as spatial relations and Human-Object Interactions (HOIs). Rather than pursuing large data scale, PhysScene focuses on distinctive characteristics of experimental scenes, including domain-specific instruments, function-driven interactions among experimental objects, high relation density, and strong semantic constraints. These properties enable systematic evaluation of scene understanding in experimental contexts. Notably, the structured annotations provided by PhysScene may also contribute to intelligent experiment monitoring, workflow analysis, and broader data-driven educational applications.\par

We further conduct a comprehensive statistical analysis of PhysScene and compare it with existing datasets to highlight its distinctive coverage and complexity. To facilitate systematic evaluation, we conduct extensive experiments under Predicate Classification (PredCls) and Scene Graph Detection (SGDet) protocols \cite{xu2017scene, chen2025data}, and report results of representative state-of-the-art models in both fully supervised and zero-shot settings. Experimental results demonstrate that PhysScene poses new challenges for relational semantic modeling and generalization to unseen objects and relations. We anticipate that PhysScene will promote future research on extending SGG methodologies to scientific experimental scenarios. \par

\begin{figure*}
	\centering
	\includegraphics[width=0.8\textwidth]{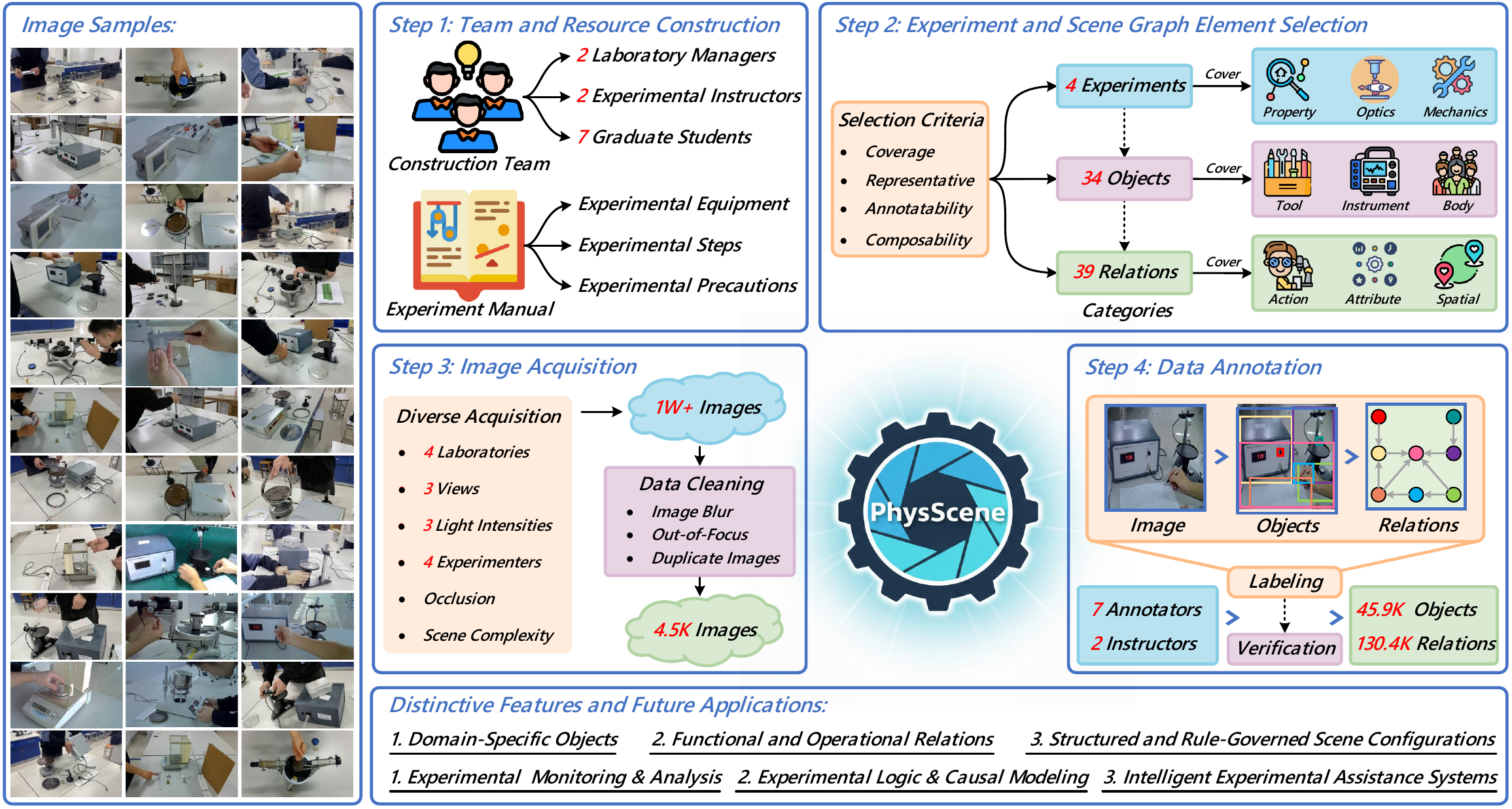}\\
	\caption{An overview of the PhysScene dataset, including the dataset construction pipeline, representative image samples, and key dataset characteristics.}
	\label{fig_dataset}
\end{figure*}

In summary, our contributions are threefold:\par

(1) We introduce PhysScene, the first SG dataset specifically designed for physics laboratory experiments. It captures domain-specific instruments, structured experimental setups, and diverse relational patterns in experimental scenarios.\par

(2) We perform a comprehensive statistical analysis of PhysScene, revealing its notable challenges, including high scene variability, complex relational structures, imbalanced distributions, and underexplored semantic constraints.\par

(3) We conduct comprehensive experiments under PredCls and SGDet protocols in both fully supervised and zero-shot settings, providing a foundation for advancing SGG research in scientific experimental domains.\par

\section{PhysScene}
\label{sec:phys}

\subsection{Data Construction}

PhysScene is constructed to capture visual scenes arising from real physics experiments. As illustrated in Figure~\ref{fig_dataset}, the dataset construction pipeline consists of four stages.

\textbf{Step 1: Team and Resource Construction.} The data collection and annotation are conducted by 7 trained graduate students under the supervision of laboratory managers and physics instructors to maintain the quality and reliability of the dataset. The dataset construction is guided by official physics experiment manuals used in university courses, which specify experimental procedures, instrument usage, and operational constraints. These manuals informed the design of object categories and relation taxonomy, ensuring that the annotations reflect real experimental operations rather than artificially curated interactions. More details can be found on our project website.

\textbf{Step 2: Experiment and SG Element Selection.} To ensure representative coverage of physics laboratory activities, we select four core experiments: Object Density Measurement, Spectrometer-Based Measurement, Surface Tension Measurement, and Rigid Body Inertia Determination. These experiments cover three common types of physics experiments, namely optics, mechanics, and measurement-related setups. Guided by experiment manuals and instructors, we defined the SG vocabulary according to three criteria: coverage of common experimental operations, inclusion of representative instruments, and composability for relational reasoning. This process results in a taxonomy of 34 object categories and 39 relation categories, including spatial relations, HOIs, and object attributes, where object attributes are represented as self-loop relations \cite{krishna2017visual, park2023viplo} to maintain compatibility with existing SGG methods and enable a unified graph representation.

\textbf{Step 3: Image Acquisition.} The acquisition process is guided by three principles to enhance data diversity and reliability.

(1) Scene diversity. PhysScene spans four representative physics experiments, encompassing a wide spectrum of experimental instruments and procedural stages. Beyond HOIs, the annotations further incorporate object attributes and inter-object spatial relations, leading to richer semantic coverage for SG modeling~\cite{liang2019vrr}.

(2) Collection variability. Images are collected across four laboratories with substantial variations in viewpoints, focal lengths, illumination, and occlusion patterns. This diversity mitigates overfitting to constrained setups and promotes generalization to real-world scenarios~\cite{gao2022review}.

(3) Standardization with flexibility. While all experiments follow standardized experimental manuals, students' execution introduces individual variations. This balance ensures consistency in experimental content while preserving natural variability in object configurations and functional relations.

Images are captured using a Newmine Q40 HD camera recording at 4K resolution (3840×2160) and 30 FPS. The high resolution enables precise capture of instrument details and spatial arrangements, while the 90° distortion-free lens provides wide coverage with minimal geometric distortion. After filtering blurred, duplicated, or uninformative frames, we retain 4.5K images for annotation. These images reflect realistic experimental conditions rather than curated demonstration scenes.

\textbf{Step 4: Data Annotation.} We develop a comprehensive annotation schema covering 34 object categories, 5 spatial relations, 10 object attributes, and 24 HOIs. Each image is annotated with relational triplets $\langle subject, relation, object \rangle$, together with object labels and bounding box coordinates \cite{chen2025data}. This structured representation explicitly captures objects and their pairwise relations, providing a foundation for SG modeling in experimental scenes \cite{chang2021comprehensive}.

Given the dense object layouts and relation-rich experimental scenes, the annotation process requires careful quality control. Seven trained annotators perform labeling using the LabelMe tool under the supervision of laboratory instructors. To improve consistency, we develop detailed annotation guidelines derived from laboratory manuals, covering object categories, attributes, and relation definitions. Annotators are assigned to specific experiment types to ensure familiarity with the corresponding instruments and experimental content.

The annotation process follows a two-stage workflow. In the first stage, annotators independently label objects and relations according to the guidelines. In the second stage, the annotations are reviewed by instructors to identify potential labeling errors or omissions. As object categories and spatial boundaries are typically well defined in experimental settings, object annotations are largely unambiguous, while fewer than 3\% of relation annotations require additional calibration. The final dataset contains 4.5K images with 45.9K annotated object instances and 130.4K relational triplets.

\begin{table*}
	\caption{Comparison of representative SG datasets across domains. Obj.: object attribute, Spa.: spatial relationship, Int.: interaction, Exp.: experimental attribute, Cls.: class, BPI: box per image, TPI: triplet per image.}
	\label{tab:statics}
    \vspace{-5pt}
	\centering
	\footnotesize
	\resizebox{0.8\textwidth}{!}{
		\begin{tabular}{c|c|c|c|cccc|ccc|ccc}
			\toprule 
			& & & & \multicolumn{4}{c|}{\textbf{Attribute}} & \multicolumn{3}{c|}{\textbf{Object}} & \multicolumn{3}{c}{\textbf{Relation}}\\
			\textbf{Dataset} & \textbf{Year} & \textbf{Domain} & \textbf{Image} & Obj. & Spa. & Int. & Exp. & Bbox & Cls. & BPI & Triplet & Cls. & TPI\\
			\midrule 
			VG150 \cite{krishna2017visual} & 2017 & Generic Natural Scene & 87.7K & \textcolor{green}{\cmark} &  \textcolor{green}{\cmark} &  \textcolor{green}{\cmark} & \textcolor{red}{\xmark} & 739.0K & 150 & 8.4 & 413.3K & 50 & 4.7\\
			SpatialVOC2K \cite{belz2018spatialvoc2k} & 2018 & Generic Natural Scene & 2.0K & \textcolor{red}{\xmark} & \textcolor{green}{\cmark} & \textcolor{red}{\xmark} & \textcolor{red}{\xmark} & 5.8K & 20 & 2.9 & 9.8K & 34 & 4.8 \\
			SpatialSense \cite{yang2019spatialsense} & 2019 & Generic Natural Scene & 11.6K & \textcolor{red}{\xmark} & \textcolor{green}{\cmark} & \textcolor{red}{\xmark} & \textcolor{red}{\xmark} & 35.0K & 3679 & 3.0 & 13.2K & 9 & 1.1\\
			VrR-VG \cite{liang2019vrr} & 2019 & Generic Natural Scene & 59.0K & \textcolor{green}{\cmark} &  \textcolor{green}{\cmark} &  \textcolor{green}{\cmark} & \textcolor{red}{\xmark} & 282.5K & 1600 & 4.8 & 203.4K & 117 & 3.5\\
			GRTRD \cite{chen2021message} & 2021 & Remote Sensing & 3.2K &\textcolor{green}{\cmark} &  \textcolor{green}{\cmark} &  \textcolor{red}{\xmark} & \textcolor{red}{\xmark} & 20.0K & 12 & 6.2 & 18.6K & 33 & 5.8\\
			PSG \cite{yang2022panoptic} & 2022 & Generic Natural Scene & 48.7K & \textcolor{green}{\cmark} &  \textcolor{green}{\cmark} &  \textcolor{green}{\cmark} & \textcolor{red}{\xmark} & 56.2K & 133 & 1.1 & 273.0K & 56 & 5.6\\
			Haystack \cite{lorenz2023haystack} & 2023 & Generic Natural Scene & 11.3K & \textcolor{green}{\cmark} &  \textcolor{green}{\cmark} & \textcolor{green}{\cmark} & \textcolor{red}{\xmark} & 145.9K & 132 & 12.8 &26.0K & 56 & 2.3\\
			AUG \cite{li2024aug}& 2024 & Urban Aerial & 0.4K & \textcolor{green}{\cmark} &  \textcolor{green}{\cmark} & \textcolor{red}{\xmark} & \textcolor{red}{\xmark} & 25.6K & 76 & 64.0 & 17.0K & 61 & 42.4\\
			STAR \cite{li2025star} & 2025 & Remote Sensing & 31.1K & \textcolor{green}{\cmark} &  \textcolor{green}{\cmark} &  \textcolor{red}{\xmark} & \textcolor{red}{\xmark} & 219K & 48 & 7.1 & 40.1K & 58 & 12.9\\ 
			\midrule
			PhysScene & 2026 & Physics Experiment & 4.5K & \textcolor{green}{\cmark} & \textcolor{green}{\cmark} & \textcolor{green}{\cmark} & \textcolor{green}{\cmark} & 45.9K & 34 & 10.2 & 130.4K & 39 & 29.0\\
			\bottomrule
		\end{tabular}
		}
        \vspace{-5pt}
\end{table*}

\subsection{Data Statistics}

Table~\ref{tab:statics} compares PhysScene with several representative SG datasets across different domains. Early datasets such as VG150 \cite{krishna2017visual}, SpatialSense \cite{yang2019spatialsense}, and PSG \cite{yang2022panoptic} primarily focus on generic natural scenes and are typically large-scale. More recent datasets, including GRTRD \cite{chen2021message}, AUG \cite{li2024aug}, and STAR \cite{li2025star}, extend SG modeling to aerial or remote-sensing imagery. They are often smaller in scale, as they prioritize domain-specific modeling over broad coverage. Nevertheless, their dataset size remains adequate to support related research. This shift highlights an increasing emphasis on specialized benchmarks tailored to specific domains. However, scientific experimental scenes remain largely unexplored in existing SG benchmarks, motivating the construction of PhysScene.

\begin{figure*}[tbp]
	\centering
	\includegraphics[width=0.72\textwidth]{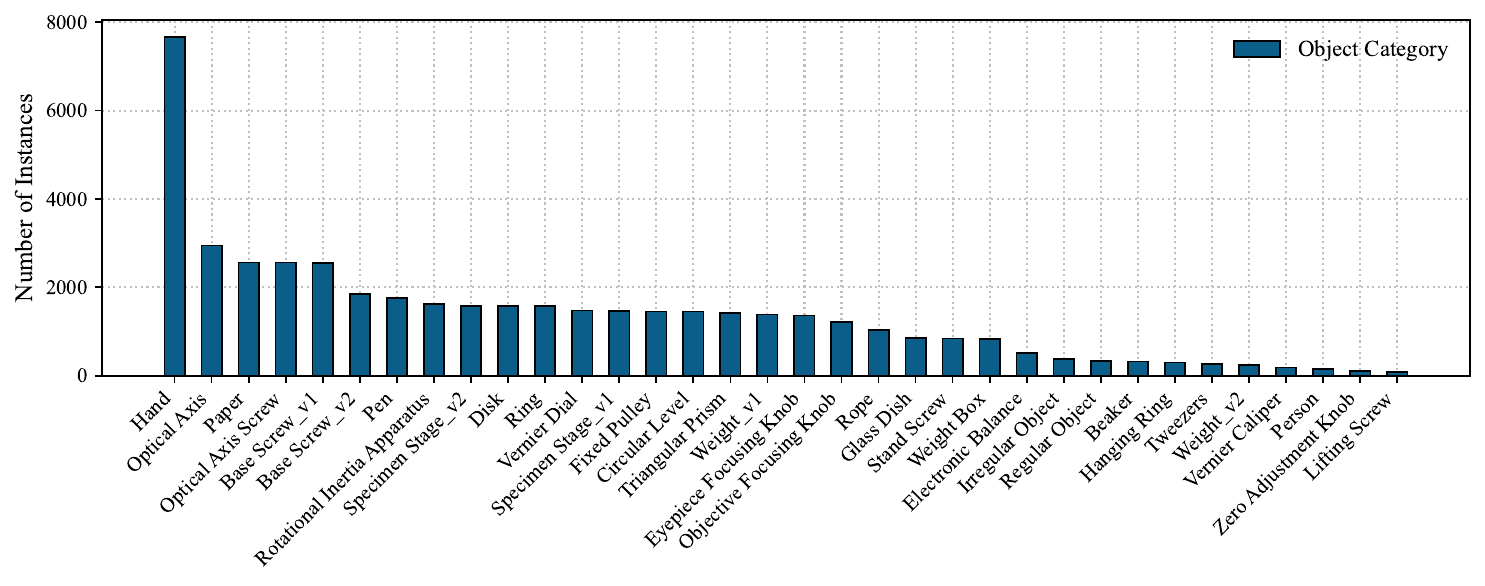}\\
	\vspace{-5pt}
	\caption{Distribution of object categories in PhysScene. The dataset covers 34 object categories spanning experimental instruments, laboratory tools, and human operators.}
	\label{fig_obj}
	\vspace{-5pt}
\end{figure*}

\begin{figure*}[tbp]
	\centering
    \vspace{-5pt}
	\scalebox{1.5}{%
		\raisebox{0pt}{\includegraphics[height=3.05cm]{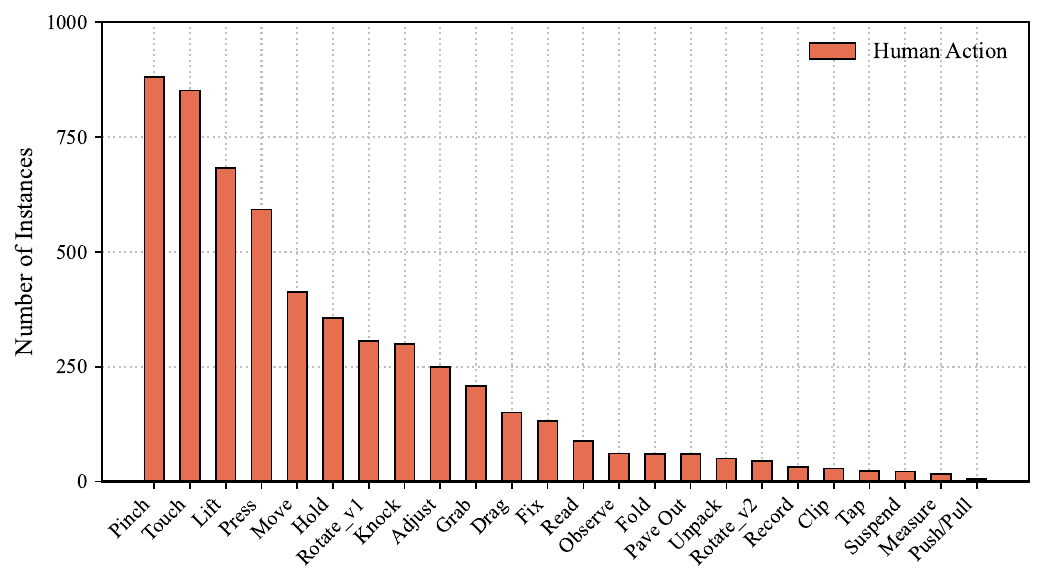}}
		\hspace{0.2cm}
		\raisebox{0pt}{\includegraphics[height=3.05cm]{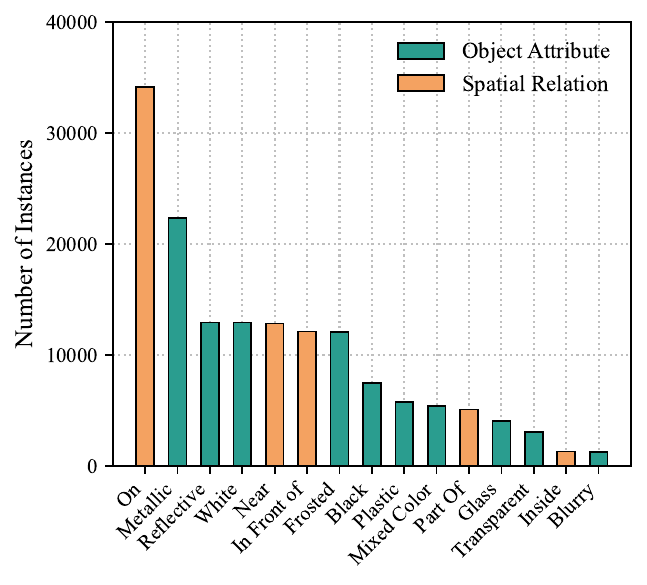}}
	}
	\vspace{-5pt}
	\caption{Distribution of relation annotations in PhysScene, including human actions, object attributes, and spatial relations.}
	\label{fig_rel}
	\vspace{-5pt}
\end{figure*}

PhysScene contains 4.5K images with 45.9K annotated object instances and 130.4K relational triplets, with an average of 10.2 objects and 29.0 relations per image. Compared with existing datasets, PhysScene provides relatively dense relational annotations while maintaining a moderate number of object and relation categories. Beyond standard annotations of attributes, spatial relations, and HOIs, PhysScene implicitly captures unique experimental characteristics such as configuration constraints, functional relationships, measurable properties, and logical dependencies, enriching the semantic representation of experimental scenes \cite{zou2025physlab}.

Figure~\ref{fig_obj} illustrates the distribution of object categories in PhysScene. The dataset covers 34 object classes spanning experimental instruments, laboratory tools, and human participants. The distribution exhibits a long-tailed pattern, where frequently used components such as optical axes and screw appear more often, while specialized instruments occur less frequently.

Figure~\ref{fig_rel} shows the distributions of three relation groups: human actions, object attributes, and spatial relations. Human actions capture common experimental operations such as touching, lifting, pressing, and holding. Object attributes describe observable properties such as material or appearance, while spatial relations characterize the arrangements between objects. Within each relation group, the frequencies of relation categories also exhibit a long-tailed distribution. Moreover, the number of annotations across these groups varies considerably. Object attributes and spatial relations appear in most images, as they describe inherent properties and spatial configurations of objects, whereas human actions are typically associated with specific experimental operations and therefore occur in fewer scenes. The imbalanced distributions of object and relation categories reflect the characteristics of real experimental scenes and introduce challenges for generalization to low-frequency categories in SGG research.

\section{Experiments}
\label{sec:exp}

\subsection{Experimental Setup}

\textbf{Benchmark Methods.} To establish comprehensive benchmarks on PhysScene, we evaluate a set of representative SGG approaches spanning multiple research paradigms, including early message-passing models \cite{xu2017scene}, global-context reasoning methods \cite{zellers2018neural}, debiasing techniques \cite{tang2020unbiased, zhang2024bridging}, as well as recent open-vocabulary SGG methods \cite{he2022towards, zhang2023learning, chen2025data}. We adopt the official implementations released by the authors and follow the training configurations reported for the VG150 dataset in the original papers to ensure fair comparison. For methods that do not provide implementations or report results under certain evaluation settings, the corresponding entries in the result tables are left unreported (marked as "-").

\textbf{Evaluation Metrics.} Following standard practice in SGG research \cite{xu2017scene, zellers2018neural}, we adopt $Recall@K$ as the primary evaluation metric, where $K \in \{50, 100\}$. $Recall@K$ measures the fraction of Ground-Truth (GT) relational triplets successfully retrieved among the top-$K$ predictions. A predicted triplet is considered correct if and only if the following conditions are satisfied:

(1) The predicted subject and object bounding boxes both achieve an Intersection over Union (IoU) greater than 0.5 with the corresponding GT boxes.\par

(2) The predicted subject and object categories match the corresponding GT labels.\par

(3) The predicted relation category matches the GT relation.\par

Formally, given the set of GT triplets $G$ and the top-$K$ predicted triplets $P_K$, $Recall$@$K$ is defined as:

\begin{equation}
	Recall\text{@}K = \frac{|G \cap P_K|}{|G|}.
\end{equation}

\textbf{Evaluation Protocols.} We evaluate models under two standard protocols \cite{xu2017scene, zellers2018neural}:\par

(1) Predicate Classification (PredCls). GT bounding boxes and categories of objects are provided, and the task is to predict relations between object pairs.\par

\begin{table*}[htbp]
	\caption{Comparison of state-of-the-art methods on the PhysScene and VG150 datasets. Results are reported as Recall@50/100.}
	\label{tab:com1}
	\small
	\centering
	\resizebox{0.9\textwidth}{!}{
		\begin{tabular}{c|c|cccc|cccc|cccc}
			\toprule
			& & \multicolumn{4}{c|}{\textbf{Cs-SGG}} & \multicolumn{4}{c|}{\textbf{OvD-SGG (70\% Base + 30\% Novel)}} & \multicolumn{4}{c}{\textbf{OvR-SGG (70\% Base + 30\% Novel)}}\\
			& & \multicolumn{2}{c}{PredCls} & \multicolumn{2}{c|}{SGDet}& \multicolumn{2}{c}{PredCls} & \multicolumn{2}{c|}{SGDet} &  \multicolumn{2}{c}{PredCls} & \multicolumn{2}{c}{SGDet}\\
			\cmidrule(lr){3-4} \cmidrule(lr){5-6} \cmidrule(lr){7-8} \cmidrule(lr){9-10} \cmidrule(lr){11-12} \cmidrule(lr){13-14}
			\textbf{Method} &\textbf{Year} & PhysScene & VG150 & PhysScene & VG150 & PhysScene & VG150 & PhysScene & VG150 & PhysScene & VG150& PhysScene & VG150\\
			\midrule
			IMP \cite{xu2017scene} & 2017 & 37.4/44.2 & 44.8/53.1 & 29.4/31.3 & 20.7/24.5 & 30.2/35.6 & 40.0/43.4 & 4.7/6.6 & 2.9/3.4 & 18.3/24.6 & 25.7/26.6 & 17.7/19.3 & 12.6/14.7\\
			Motifs \cite{zellers2018neural} & 2018 & 59.6/62.3 & 65.2/67.1 & 47.3/49.8 & 32.1/36.9 & 33.7/36.0 & 41.1/44.7 & 4.9/8.2 & 3.4/3.9 & 25.5/31.2 & 27.1/28.0 & 21.6/22.4 & 15.4/17.0\\
			Motifs+TDE \cite{tang2020unbiased} & 2020 & 49.8/57.2 & 50.8/55.8 & 36.7/39.5 & 16.9/20.3 & 36.9/42.1 & 38.3/40.4 & 5.1/8.6 & 3.5/4.1 & 32.4/33.7 & 29.0/29.8 & 23.8/26.6 & 15.5/17.4\\
			SVRP \cite{he2022towards} & 2022 & 61.4/62.0 & 60.2/62.3 & 44.6/49.2 & 31.8/35.8 & - & 47.6/49.9 & - & - & - & - & - & -\\
			VS$^3$ \cite{zhang2023learning} & 2023 & 62.9/64.4 & - & 50.6/55.9 & 36.6/41.5 & 46.7/48.8 & 55.9/58.2 & 26.1/30.3 & 23.1/28.5 & 33.1/34.7 & 24.3/25.0 & 22.5/24.7 & 15.6/17.3\\
			OpenPSG \cite{zhou2024openpsg} & 2024 & 60.7/63.3 & 60.2/71.4 & 38.4/49.3 & 20.3/23.6 & - & - & - & - & - & - & - & -\\
			Motifs+VTSCN \cite{zhang2024bridging} & 2024 & 42.3/45.8 & 41.3/43.5 & 33.1/38.0 & 17.1/21.0 & - & - & - & - & - & - & - & -\\
			OvSGTR \cite{chen2025data} & 2025 & 64.8/65.5 & - & 49.6/51.6 & 36.4/42.4 & 54.6/56.0 & 60.8/62.3 & 25.7/28.9 & 21.4/26.2 & 43.3/47.8 & 41.0/44.7 & 31.2/34.1 & 22.9/26.7\\
			\bottomrule
		\end{tabular}
	}

\end{table*}

\begin{table*}[htbp]
	\caption{Comparison of state-of-the-art methods on novel classes of PhysScene and VG150 datasets.}
    \vspace{-5pt}
	\label{tab:com2}
	\footnotesize
	\centering
    \resizebox{0.7\textwidth}{!}{
	\begin{tabular}{c|c|cccc|cccc}
		\toprule
		& &  \multicolumn{4}{c|}{\textbf{OvD-SGG (30\% Novel)}} & \multicolumn{4}{c}{\textbf{OvR-SGG (30\% Novel)}}\\
		& & \multicolumn{2}{c}{PredCls} & \multicolumn{2}{c|}{SGDet} &  \multicolumn{2}{c}{PredCls} & \multicolumn{2}{c}{SGDet}\\
		\cmidrule(lr){3-4} \cmidrule(lr){5-6} \cmidrule(lr){7-8} \cmidrule(lr){9-10}
		\textbf{Method} &\textbf{Year} & PhysScene & VG150 & PhysScene & VG150 & PhysScene & VG150& PhysScene & VG150\\
		\midrule
		IMP \cite{xu2017scene} & 2017 & 26.4/29.9 & 37.0/39.5 & - & - & - & - & - & -\\
		Motifs \cite{zellers2018neural} & 2018 & 27.1/31.4 & 39.5/41.1 & - & - & - & - & - & -\\
		Motifs+TDE \cite{tang2020unbiased} & 2020 & 30.0/32.8 & 34.2/36.4 & - & - & - & - & - & - \\
		VS$^3$ \cite{zhang2023learning} & 2023 & 39.9/43.2 & 46.9/49.1 & 18.9/22.0 & 10.1/13.7 & - & - & - & - \\
		OvSGTR \cite{chen2025data} & 2025 & 48.8/53.3 & 59.3/61.0 & 17.4/21.3 & 15.6/20.0 & 32.4/36.9 & 32.9/37.9 & 27.2/29.0 & 16.4/19.7\\
		\bottomrule
	\end{tabular}}
        \vspace{-5pt}
\end{table*}

(2) Scene Graph Detection (SGDet). Both object detection and relation prediction are performed jointly, requiring models to localize objects and infer their pairwise relations.\par

\textbf{Supervision Settings.} To further evaluate model generalization, we adopt three supervision settings following \cite{chen2025data}:\par

(1) Closed-set SGG (Cs-SGG). Both object and relation vocabularies are restricted to the categories observed during training, consistent with the conventional setting of VG150.\par

(2) Open-vocabulary Detection-based SGG (OvD-SGG). Models are required to recognize novel object categories. Specifically, 70\% of object categories are randomly selected as base categories and used for training. The remaining 30\% are treated as novel categories and excluded from training. During evaluation, the test set contains both base and novel categories, and the model is expected to recognize objects from both groups.\par

(3) Open-vocabulary Relation-based SGG (OvR-SGG). It follows the same setting as OvD-SGG, except that the 70\%/30\% split is applied to relation categories rather than object categories, while the object vocabulary is shared across training and testing.\par

\subsection{Experimental Results}

We establish benchmark results on PhysScene and compare them with those on the widely used VG150 dataset as a reference.

As shown in Table~\ref{tab:com1}, under the Cs-SGG setting, the average $Recall@50/100$ of the six shared models is 51.9/55.8 on PhysScene and 53.6/58.9 on VG150 under PredCls. Under SGDet, the corresponding values are 38.2/42.8 on PhysScene and 23.2/27.0 on VG150. These results indicate that PhysScene is more challenging than VG150 under the PredCls protocol, while relatively less challenging under SGDet. This contrast reveals a key difference between the two datasets: PhysScene primarily challenges models in relation prediction, whereas VG150 places greater emphasis on object detection. This observation reflects the unique characteristics of PhysScene, where the functional semantics of instruments and task-driven interactions make relation classification particularly challenging.

When introducing 30\% novel object categories, the relative performance difference (measured as the percentage increase between the maximum and minimum model performance) in $Recall@50$ among models under the PredCls protocol reaches 80.8\% on PhysScene. In comparison, the corresponding difference on VG150 is 59.5\%. Under the SGDet protocol, the corresponding performance gap increases to 455.3\% for PhysScene and 696.6\% for VG150.  When introducing 30\% novel relation categories, the relative sensitivities to object detection and relation prediction across the two datasets follow a similar trend to those observed under the Cs-SGG and OvD-SGG settings. These findings indicate that PhysScene provides a more discriminative evaluation benchmark, where relation prediction performance gaps across models are more pronounced.

Although relatively few existing methods report results under the OvD-SGG and OvR-SGG settings with independently constructed novel classes, we nevertheless establish these benchmarks on PhysScene to facilitate future research on open-vocabulary SGG. Comparing Table~\ref{tab:com1} and Table~\ref{tab:com2} shows that performance on novel categories consistently lags behind that on the corresponding base categories, indicating the challenge of generalizing to unseen objects and relations. This observation highlights the need for models with stronger robustness and generalization ability \cite{zhang2023learning} to better adapt to open-world experimental scenes.

\section{Limitations and Future Work}
\textbf{Dataset Scale and Diversity.}
While PhysScene is comparable in scale to existing domain-specific SG datasets, it is smaller than large-scale general-purpose datasets. In addition, the current release focuses on four representative physics experiments, with opportunities to further expand the diversity of experimental scenes. To this end, we are actively collecting additional images from six physics experiments, as well as chemistry and biology laboratory setups. The expanded dataset will be released through our open-source repository. We are also extending the annotations with image captions to support cross-modal tasks in experimental scenes, such as visual question answering.

\textbf{Procedural Modeling.}
In the current version of PhysScene, each image is annotated as a static SG. While these annotations capture detailed scene information, they do not explicitly model the semantic dependencies across different experimental steps. Future work will explore extending image-level SGs to video-level representations, enabling procedural understanding and contextual modeling of experimental workflows.

\textbf{Detailed Fine-grained Evaluation.}
PhysScene includes images collected under diverse laboratory conditions, with variations in illumination, viewpoints, occlusion, and scene complexity, as well as a wide range of relation types, including attributes, spatial arrangements, and interactions. However, the current evaluation primarily reports overall performance without explicitly analyzing results across different conditions or relation subsets. Due to space constraints, fine-grained evaluations are not presented in this paper. We plan to release additional experimental results and analyses through the project website, including benchmarks that explicitly account for different conditions and subsets.

\textbf{Downstream Applications.}
Current experiments focus on standard SGG benchmarks to ensure fair comparison across datasets and methods. However, the broader applicability of PhysScene to downstream tasks remains to be explored. Future work will investigate a wider range of multimodal applications enabled by PhysScene, including instruction generation, graph-to-image generation, and visual question answering. The structured annotations provided by PhysScene also offer promising opportunities for developing multimodal tutoring systems and intelligent experimental assistants.

\section{Conclusion}

In this paper, we introduce PhysScene, the first SG dataset specifically designed for physics laboratory experiments. PhysScene captures domain-specific instruments, structured experimental setups, and functional relations that are largely absent from existing datasets. Moreover, PhysScene is characterized by high relation density, diverse imaging conditions, imbalanced sample distributions, and strong semantic constraints, posing new challenges and opportunities for improving SGG algorithms. We conduct comprehensive experiments under multiple evaluation protocols and supervision settings, establishing PhysScene as a new benchmark for the SGG community and enabling future research on structured visual reasoning in scientific experimental scenes. We expect PhysScene to offer potential for advancing scene understanding and intelligent systems in scientific and educational domains.

\begin{acks}
	This work was supported in part by the Shandong Postdoctoral Science Foundation [SDZZ-ZR-202501503], National Science and Technology Major Project of China [2022ZD0119501], National Natural Science Foundation of China [52374221, 52574256], and Taishan Scholar Program of Shandong Province [TSTP20250506].
\end{acks}

\bibliographystyle{ACM-Reference-Format}
\balance
\bibliography{sample-base}

\end{document}